\documentclass[10pt]{article}
\pdfoutput=1 

\usepackage[preprint]{tmlr}
\usepackage{
  rawfonts,
  amsmath,
  amsfonts,
  pgf,
  caption,
  subfig,
  makecell,
  lastpage,
  algorithm,
  algpseudocode
}
\usepackage[hidelinks]{hyperref}
\usepackage{url}

\DeclareUnicodeCharacter{2212}{-}

\begin{document}
  \title{Online Arbitrary Shaped Clustering through Correlated Gaussian Functions}

  \author{\name Ole Christian Eidheim \email ole.c.eidheim@ntnu.no\\ \addr Norwegian University of Science and Technology,\\ Department of Computer Science,\\ NO-7491 Trondheim, Norway}

  \maketitle
  \begin{abstract}
    There is no convincing evidence that backpropagation is a biologically plausible mechanism, and
    further studies of alternative learning methods are needed. A novel online clustering algorithm is
    presented that can produce arbitrary shaped clusters from inputs in an unsupervised manner, and
    requires no prior knowledge of the number of clusters in the input data. This is achieved by finding
    correlated outputs from functions that capture commonly occurring input patterns. The algorithm
    can be deemed more biologically plausible than model optimization through backpropagation, although
    practical applicability may require additional research. However, the method yields satisfactory
    results\footnote{The source code to reproduce the results is available at https://gitlab.com/eidheim/online-arbitrary-shaped-clustering.}
    on several toy datasets on a noteworthy range of hyperparameters.
  \end{abstract}

  \section{Introduction}
  Backpropagation is not considered biologically plausible \citep{bengio2016biologically,hinton2022forward},
  and popular artificial neuron models are oversimplified compared to cortical neurons \citep{beniaguev2021single}.
  Further evidence is the relatively low firing rate in biological neurons \citep{wang2016firing},
  which limits the depth of more biologically plausible neural architectures compared to today’s
  deep artificial networks.

  In this article, a novel online clustering algorithm is presented that can produce arbitrary shaped
  clusters by finding output correlations, which exceed a given threshold, between Gaussian functions
  representing the input data. Inherently, the procedure is unsupervised and does not depend on a predetermined
  cluster count. The method is shown to produce acceptable results when applied on toy datasets, and
  is not over-reliant on the choice of hyperparameters.

  Online clustering of arbitrary shaped clusters was first introduced in \citet{hyde2015new}, and extended
  in \citet{hyde2017fully,roa2019dyclee,islam2019buffer,tareq2020online,tareq2021evolving}. These methods,
  however, are dependent on a large number of input representation points, i.e. points capturing commonly
  occurring input patterns, that must lie sufficiently close together in order for the points to be
  grouped into clusters. In comparison, the presented algorithm requires fewer such representation
  points, and makes use of separated Gaussian functions generated by the online learning rule from \citet{eidheim2022revisiting}.

  \section{Online Arbitrary Shaped Clustering}
  Let $\boldsymbol{x}\in \mathbb{R}^{D}$ be an input, and $\boldsymbol{\mu}_{i}\in \mathbb{R}^{D}$ and
  $\sigma_{i}\in \mathbb{R}_{>0}$ be center and width, respectively, of the $i$-th Gaussian function,
  where $i \in \{1,\ldots,K\}$ and $K$ is the upper bound on the number of Gaussian functions
  capturing the input patterns. The output of the $i$-th Gaussian function is then defined as:

  \[
    f_{i}(\boldsymbol{x}) = e^{-{\lVert \boldsymbol{x} - \boldsymbol{\mu}_i \rVert}_2^2 / \sigma_i}
  \]

  \noindent
  where ${\lVert \cdot \rVert}_{2}$ denotes the $l_{2}$-norm. The learning rule from \citet{eidheim2022revisiting},
  in the case of $\sigma_{i}= \sigma \  \forall i \in \{1, \ldots, K \}$, is used to find the
  Gaussian functions:

  \[
    \Delta \boldsymbol{\mu}_{i}= \frac{\eta}{\sigma}\left ( f_{i}(\boldsymbol{x})(\boldsymbol{x}-\boldsymbol{\mu}_{i}) - 2 \lambda \sum_{j \ne i}f_{i}(\boldsymbol{\mu}_{j}) (\boldsymbol{\mu}_{j}- \boldsymbol{\mu}_{i}) \right )
  \]

  \noindent
  where $\lambda \in \mathbb{R}_{>0}$ controls how similar the Gaussian functions can be, and
  $\eta\in \mathbb{R}_{>0}$ is the learning rate. Lowering $\lambda$ and $\sigma$ has the same effect,
  that is potentially more closely positioned Gaussian functions, but a small value of $\sigma$ may lead
  to insufficient attraction of the Gaussian functions toward the input.

  Additionally, for each $\boldsymbol{x}$, a scatter-like matrix $\boldsymbol{Q}\in \mathbb{R}^{K
  \times
  K}$ is updated as follows:

  \[
    \Delta Q_{k, l}=\frac{f_{k}(\boldsymbol{x}) f_{l}(\boldsymbol{x})}{{\lVert \boldsymbol{f}_{1:K}(\boldsymbol{x}) \rVert}_{p}^{2}}, \quad Q_{k,l}(0)=0
  \]

  \noindent
  where $k,l \in \{1,\ldots,K\}$. In the experiments in Section \ref{results}, the $l_{\infty}$-norm
  was used, but other values of $p$ are discussed.

  \subsection{Assigning Cluster Labels}
  Cluster labels for the Gaussian functions can be obtained after any $N$ inputs from the uncentered
  sample Pearson correlation coefficient $\boldsymbol{R}\in (0,1]^{K \times K}$ derived from $\boldsymbol
  {Q}$:

  \[
    R_{k, l}=\frac{Q_{k, l}}{\sqrt{Q_{k, k}} \sqrt{Q_{l, l}}}
  \]

  \noindent
  The cluster labels $\boldsymbol{y}_{1:K}\in \{0, \ldots , L\}^{K}$ of the Gaussian functions are
  then assigned by the following algorithm:

  \begin{algorithm}
    \caption{Assigning cluster labels $\boldsymbol{y}_{1:K}$ to the Gaussian functions $\boldsymbol{f}
    _{1:K}$ from the uncentered sample Pearson correlation coefficient $\boldsymbol{R}$.}
    \label{algorithm1}
    \begin{algorithmic}
      \State $\boldsymbol{y}_{1:K}\gets (0, \ldots, 0)$

      \State $L \gets 0$

      \Statex

      \Function{Assign}{$k$}

      \State $y_{k}\gets L$

      \For{$l \gets 1 \text{ to }K$}

      \If{$y_{l}= 0 \textbf{ and }R_{k,l}> \tau$}

      \State $\Call{Assign}{l}$

      \EndIf

      \EndFor

      \EndFunction

      \Statex

      \For{$k \gets 1 \text{ to }K$}

      \If{$y_{k}= 0$}

      \State $L \gets L+1$

      \State $\Call{Assign}{k}$

      \EndIf

      \EndFor
    \end{algorithmic}
  \end{algorithm}

  \noindent
  where $L$ is the number of separate clusters found, $\tau\in (0,1]$ is the correlation threshold, and
  $y_{i}= 0$ indicates that the $i$-th Gaussian function is not yet assigned to any cluster label.

  In the experiments in Section \ref{results}, the correlation threshold was set to
  $\tau=\frac{1}{9}$, which given $K=2$, $N=5$ and $p=\infty$, corresponds to the value of $R_{1,2}$
  from the Gaussian function outputs $(f_{1}(\boldsymbol{x}^{(1)}), \ldots, f_{1}(\boldsymbol{x}^{(N)}
  )) \approx (1, 1, \frac{1}{2}, 0, 0)$ and $(f_{2}(\boldsymbol{x}^{(1)}), \ldots, f_{2}(\boldsymbol{x}
  ^{(N)})) \approx (0, 0, \frac{1}{2}, 1, 1)$. The threshold $\tau$ can in this way be derived
  independently of the dimensions $D$ of the input domain.

  \section{Results and Discussion}
  \label{results}

  \begin{figure}[tb]
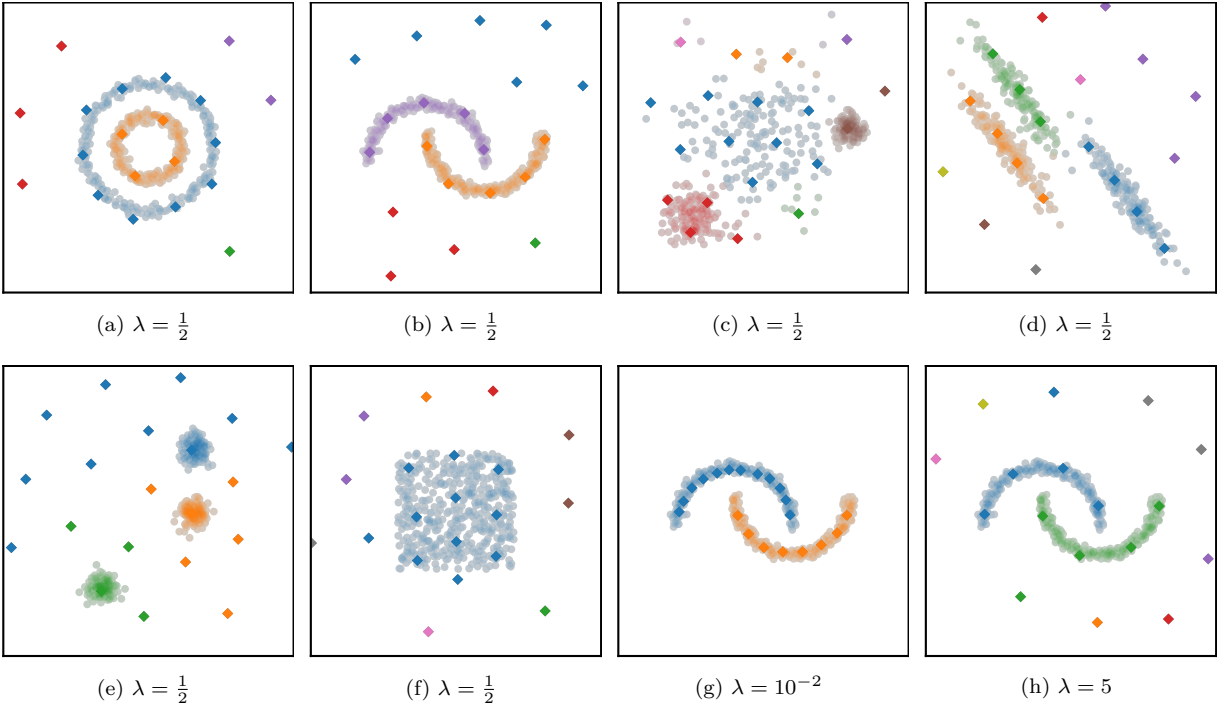

    \setlength{\tabcolsep}{1.5pt}
    \begin{tabular}{lllll}
      \subfloat[][\label{1a}$\lambda=\frac{1}{2}$]{\input{figure1a.pgf}} & \subfloat[][\label{1b}$\lambda=\frac{1}{2}$]{\input{figure1b.pgf}} & \subfloat[][\label{1c}$\lambda=\frac{1}{2}$]{\input{figure1c.pgf}} & \subfloat[][\label{1d}$\lambda=\frac{1}{2}$]{\input{figure1d.pgf}} \\
      \subfloat[][\label{1e}$\lambda=\frac{1}{2}$]{\input{figure1e.pgf}} & \subfloat[][\label{1f}$\lambda=\frac{1}{2}$]{\input{figure1f.pgf}} & \subfloat[][\label{1g}$\lambda=10^{-2}$]{\input{figure1g.pgf}}     & \subfloat[][\label{1h}$\lambda=5$]{\input{figure1h.pgf}}
    \end{tabular}
    \caption{The presented method applied on the datasets from the scikit-learn clustering comparison \citep{scikit-learn}. The diamond markers show the locations of the Gaussian function centers $\boldsymbol
    {\mu}_{1:K}$, while the transparent circles represents a subset of the inputs. The color of a marker indicates the label of the given or closest Gaussian center, but the specific color has no importance.}
    \label{figure1}
  \end{figure}

  Example results from the presented method can be seen in Figure \ref{figure1}. The datasets used are
  from the scikit-learn clustering comparison \citep{scikit-learn}, and the datasets in Subfigures \ref{1a}
  and \ref{1c} to \ref{1f} were scaled by a factor of $\frac{11}{10}$, $\frac{1}{4}$, $\frac{1}{2}$,
  $\frac{1}{8}$ and $2$, respectively. For all experiments, $D =2$, $K =20$, $N =10^{5}$,
  $\sigma=10^{-1}$, $\eta=2 \times 10^{-2}$, $p=\infty$, and $\tau=\frac{1}{9}$. Each Gaussian
  function center was initialized from a uniform distribution over $[-\frac{1}{2}, \frac{1}{2})^2$.

  Subfigures \ref{1a} to \ref{1f} show acceptable clustering results even though identical
  hyperparameters were used. Furthermore, the experiments in Subfigures \ref{1b}, \ref{1g} and \ref{1h}
  achieved similar clustering results with notably different $\lambda$ values. The method thus
  proves to be somewhat robust with respect to the most substantial hyperparameters in \citet{eidheim2022revisiting},
  namely $\lambda$ and $\sigma$.

  Different values of $p$, as well as no normalization, were tested on the datasets used to create
  Subfigures \ref{1a} and \ref{1b}. The values of $\tau$ that produced satisfactory results are shown
  in Table \ref{table1}, where $\tau$ was evaluated in a range with step size of $0.01$. The remaining
  hyperparameters were equal to those used in the aforementioned subfigures. Evidently, $p= \infty$ produced
  the largest sets of acceptable thresholds $\tau$.

  \begin{table}[h!]
    \centering
    \begin{tabular}{l!{\vrule width 2\arrayrulewidth}l|l}
      \Xhline{2\arrayrulewidth}                                                                                                    & Dataset in Subfigure \ref{1a}          & Dataset in Subfigure \ref{1b}          \\
      \Xhline{2\arrayrulewidth} \makecell[c]{No normalization, i.e.\\$\Delta Q_{k, l}=f_{k}(\boldsymbol{x})f_{l}(\boldsymbol{x})$} & $\tau\in \{0.04\}$                     & $\tau\in \{0.02, 0.03, \ldots, 0.09\}$ \\
      \hline
      \hspace{1.15cm} $p=\frac{1}{2}$                                                                                              & $\tau\in \{0.04, 0.05\}$               & $\tau\in \{0.02, 0.03, \ldots, 0.09\}$ \\
      \hline
      \hspace{1.15cm} $p=1$                                                                                                        & $\tau\in \{0.05, 0.06, \ldots, 0.09\}$ & $\tau\in \{0.03, 0.04, \ldots, 0.13\}$ \\
      \hline
      \hspace{1.15cm} $p=2$                                                                                                        & $\tau\in \{0.07, 0.08, \ldots, 0.12\}$ & $\tau\in \{0.03, 0.04, \ldots, 0.16\}$ \\
      \hline
      \hspace{1.15cm} $p=4$                                                                                                        & $\tau\in \{0.08, 0.09, \ldots, 0.13\}$ & $\tau\in \{0.03, 0.04, \ldots, 0.17\}$ \\
      \hline
      \hspace{1.15cm} $p=\infty$                                                                                                   & $\tau\in \{0.08, 0.09, \ldots, 0.14\}$ & $\tau\in \{0.03, 0.04, \ldots, 0.18\}$ \\
      \Xhline{2\arrayrulewidth}
    \end{tabular}
    \caption{The thresholds $\tau$ that produced desired results on the datasets used in Subfigures \ref{1a} and \ref{1b}.}
    \label{table1}
  \end{table}

  In comparison to the work in \cite{eidheim2022revisiting}, the presented method produces more complex
  artificial neurons, one for each cluster found in Algorithm \ref{algorithm1}. Each neuron can then
  represent arbitrary shaped regions in the input domain, which is more in line with what cortical
  neurons can plausibly represent. However, each cortical neuron might also capture multiple such
  regions, which will require extensions to the learning method beyond basic clustering approaches.

  \section{Conclusion and Future Work}
  A novel online clustering algorithm is introduced that can produce arbitrary shaped clusters, and
  the procedure differs significantly from previously published online arbitrary shaped clustering
  techniques that typically require a larger storage capacity. The results show that the outlined method
  can produce satisfactory clusters in toy datasets on a notable range of hyperparameters. Although
  the method is not demonstrated on a multi-layered architecture, the preliminary results are promising
  and further research in this direction seems worthwhile.

  The algorithm is presented in its most basic form for the sake of simplicity, but the method can
  be extended in various ways to achieve improved results on more challenging datasets. For example,
  the Gaussian function centers and the matrix $\boldsymbol{Q}$ are currently updated for each input,
  but it might be advantageous to update the Gaussian function centers solely for some number of inputs
  first. Moreover, in the case of input distribution shifts, a dampening scheme could be applied to
  the Gaussian functions that have captured a substantial amount of previous input data, and conversely
  employ a progressive strategy to update $\boldsymbol{Q}$ for improved adaptation. Finally, Gaussian
  functions with negligible output can be discarded, for instance to reduce resource usage.

  \bibliography{article}
  \bibliographystyle{tmlr}
\end{document}